%% file: main.tex
\documentclass{llncs}
\usepackage[T1]{fontenc}
\usepackage{orcidlink}
\usepackage{fancybox}
\usepackage{graphicx}
\usepackage[most]{tcolorbox}
\usepackage{xcolor}
\begin{document}
%
\title{CoPlan: A Trustworthy Co-Intelligence Interface for Care Planning through Role-Based Contestable Argument Graphs}

\titlerunning{Contestable Co-Intelligence Care Planning Interface}
%
\author{Hung Truong Thanh Nguyen\inst{1,2,*}\orcidlink{0000-0002-6750-9536} \and
Hélène Fournier\inst{2}\orcidlink{0000-0002-8470-3226} \and
Piper Jackson\inst{3}\orcidlink{0000-0002-7025-5063} \and
Makoto Itoh\inst{4}\and
Shannon Freeman\inst{5} \and
Rene Richard\inst{2}\orcidlink{0000-0002-1342-6225} \and
Hung Cao\inst{1}\orcidlink{0000-0002-0788-4377}
}
\authorrunning{Hung Nguyen et al.}
%
\institute{Analytics Everywhere Lab, University of New Brunswick, Canada \and
National Research Council Canada, Canada \and
Thompson Rivers University, Canada \and
ISB Corporation, Japan \and
University of Northern British Columbia, Canada
\\
$^*$\email{hung.ntt@unb.ca}}
\maketitle              
\begin{abstract}
AI-supported care planning can help clinicians, patients, caregivers, and care teams coordinate complex decisions across clinical, functional, psychosocial, and environmental needs. However, many AI systems present recommendations as fixed outputs, limiting stakeholders’ ability to inspect, challenge, and revise plans when they conflict with clinical judgment, patient values, or real-world feasibility.
We present \textit{CoPlan -- a Co-Intelligent and Contestable Interface for Human-AI Care Planning}. CoPlan uses a multi-agent workflow in which specialized AI agents generate candidate interventions and supporting or challenging arguments, while human care planners can accept, reject, modify, or add arguments before final plan generation. Through this design, CoPlan combines co-intelligence, in which humans and AI agents contribute complementary expertise, with contestability, where recommendations remain open to inspection, revision, and justification.
We demonstrate CoPlan in an aging-in-place care planning scenario. The system supports adaptive care team recruitment, role-based argument review, final care plan generation, and practical follow-up through scheduling agents. This work contributes a contestable care planning interface and a design framing for trustworthy human-AI care planning that preserves human agency and clinical accountability.
\end{abstract}

\input{sec/1_intro}

\input{sec/2_rw}

\input{sec/3_fw}

\input{sec/4_res}

\input{sec/4_disc}
\input{sec/5_conc}

\input{sec/6_ack}

\bibliographystyle{splncs04}
\bibliography{references}

\end{document}

%% file: sec/1_intro.tex
\section{Introduction}
Care planning is a collaborative and ongoing process in which clinicians, patients, caregivers, and care teams define goals, coordinate interventions, and revise decisions as conditions change \cite{amir2013collaborative}. Rather than a fixed document, a care plan is a shared object that connects clinical evidence with patient preferences, social context, practical constraints, and professional responsibilities. This is especially important for chronic and complex conditions, where care often spans multiple providers and settings, and where planning quality depends on whether stakeholders can understand, negotiate, and update the plan together \cite{stewart2015ecology}.

Artificial intelligence (AI) systems offer new opportunities to support care planning by summarizing patient information, identifying risks, comparing interventions, and generating recommendations. Multi-agent systems (MAS) provide a promising way to organize algorithmic care planning because they can distribute reasoning across different forms of expertise, including diagnosis, medication review, rehabilitation, social support, and patient preference elicitation \cite{tang2024medagents,kim2024mdagents,amir2013collaborative,hong2024argmed}. Yet care planning cannot be reduced to the production of an optimized recommendation, since a plan that is clinically reasonable may still be unsuitable when it overlooks local workflows, patient values, caregiver capacity, or contextual information absent from the data.

This raises a key challenge for designing trustworthy AI-supported care planning interfaces. Trustworthiness cannot rely only on accuracy, model performance, or explanation, because users also need meaningful ways to respond when AI recommendations are incomplete, unsafe, biased, or misaligned with clinical judgment and patient goals \cite{act2024eu,regulation2016regulation}. While explainable AI (XAI) can help users understand why a recommendation was produced, explanation alone does not ensure that users can correct the system, challenge its assumptions, or reshape the proposed plan. Therefore, interface design becomes a critical part of trustworthy care planning, because it shapes how users encounter AI recommendations, how they examine the reasoning behind them, and how they act when the recommendation does not fit the care context \cite{nguyen2026heart2mind,ploug_four_2020,alfrink_contestable_2023}.

In this paper, we present \textit{CoPlan – a Co-Intelligent and Contestable Interface for Human-AI Care Planning}. CoPlan is designed around two principles: \textit{co-intelligence} and \textit{contestability}. \textit{Co-intelligence} refers to a mode of human-AI collaboration in which clinicians, patients, caregivers, and AI agents contribute different forms of knowledge to the care planning process. Rather than treating AI as an autonomous planner, CoPlan positions AI as a collaborative participant that helps synthesize clinical evidence, surface risks, generate alternatives, and support reflection across stakeholders.
Meanwhile, \textit{contestability} refers to the ability of stakeholders to inspect an AI-supported care recommendation, question its evidence and assumptions, provide counterevidence, express disagreement, and trigger revision or justified confirmation \cite{leofante2024contestable,dignum2025contesting,alfrink_contestable_2023}. 

Through this framing, CoPlan treats AI recommendations not as fixed decisions, but as collaborative proposals open to negotiation. This is important because trustworthy care planning requires both distributed intelligence across human and AI agents, and concrete mechanisms for challenging recommendations when they conflict with clinical judgment, patient values, or real-world feasibility.
Overall, our main contribution can be summarized as follows:
\begin{enumerate}
    \item \textbf{CoPlan, a co-intelligent and contestable interface} that makes AI-supported care recommendations inspectable, challengeable, revisable, and auditable. Our implementation is available at \url{https://github.com/Analytics-Everywhere-Lab/CAIAiPCP/}.
    \item \textbf{A contestability-centered framing} that connects trustworthy AI, human-centered design, and clinical accountability in human-AI care planning.
    \item \textbf{Design implications for future AI-supported care planning systems} that aim to preserve human agency while benefiting from AI reasoning and multi-agent coordination.
\end{enumerate}

%% file: sec/2_rw.tex
\section{Related Works}
\subsection{Human-Centered and Trustworthy AI for Care Planning}
Care planning is a collaborative healthcare activity that requires clinicians, patients, caregivers, and care teams to align clinical evidence with patient goals, social context, available resources, and practical constraints. Prior work on AI in healthcare has shown the potential of computational systems to support clinical reasoning, risk assessment, diagnosis, and treatment planning \cite{kim2024mdagents,borkowski2025multiagent}, while also emphasizing that AI should augment rather than replace human judgment \cite{topol2019high,world2024ethics}. This perspective is especially important in care planning, where decisions are not only technical or clinical, but also relational, contextual, and value-sensitive.

Human-centered AI research argues that AI systems should be designed around human goals, capabilities, responsibilities, and contexts of use \cite{shneiderman2020human,riedl2019human,nguyen2026heart2mind}. In healthcare, this means that AI-supported systems must fit clinical workflows, communicate uncertainty, support user control, and preserve accountability. Guidelines for human-AI interaction similarly emphasize that systems should make their capabilities visible, support correction, and enable users to understand and control AI behavior \cite{amershi2019guidelines,nguyen_motion2meaning_2025}. These principles are directly relevant to care planning because different stakeholders contribute different forms of knowledge: clinicians bring clinical expertise, patients bring lived experience and preferences, and caregivers bring practical knowledge about daily feasibility.

Recent work on multi-agent systems (MAS) extends this direction by showing how multiple AI agents can coordinate different forms of expertise, such as diagnosis, medication review, rehabilitation, and patient preference elicitation \cite{li2024agent,kim2024mdagents,borkowski2025multiagent,tang2024medagents}. This is useful for care planning because care work is already distributed across roles and specialties. However, existing AI and multi-agent healthcare systems often focus on recommendation quality, agent coordination, or reasoning performance, while giving less attention to the interface through which human users inspect, negotiate, and revise AI-generated plans. 

CoPlan addresses this gap by framing care planning as a \textit{co-intelligent human-AI process}, where human stakeholders and AI agents jointly contribute to planning while the interface preserves human agency and accountability.

\subsection{From Explainability to Contestable Human-AI Interfaces}
Explainable AI (XAI) has become a major approach for improving trust in AI-supported decision-making, especially in high-stakes domains such as healthcare. Explanations can help users understand what factors shaped a recommendation, why a system produced a certain output, and when the output may require further review \cite{gunning_xaiexplainable_2019,nguyen_evaluation_2021}. However, explanation alone is limited because understanding an AI recommendation does not necessarily give users a meaningful way to challenge or change it. In care planning, this limitation is critical, where a recommendation may be explainable but still inappropriate if it conflicts with patient preferences, caregiver capacity, cost, access, or clinical judgment \cite{procter2023holding}.

\begin{figure}[htbp]
    \centering
    \includegraphics[width=.9\linewidth]{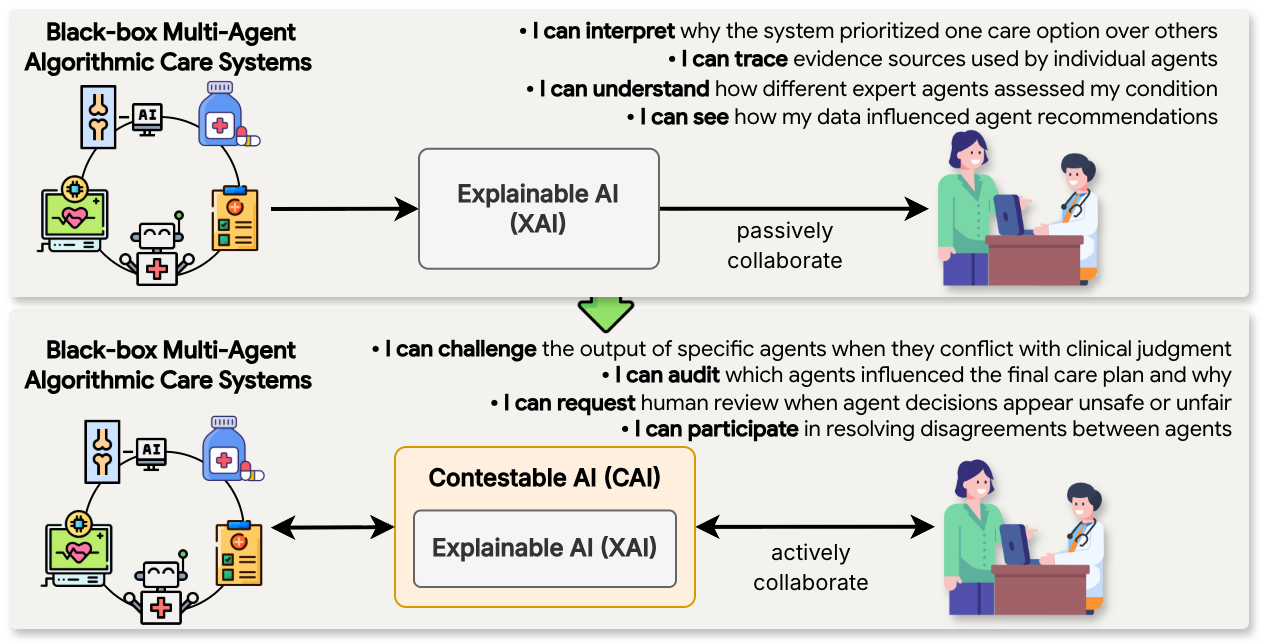}
    \caption{Human-AI Co-Intelligence Mechanism through Contestability for Agentic AI in Care Planning Systems.}
    \label{fig:fw}
\end{figure}

Trustworthy AI frameworks emphasize transparency, fairness, robustness, accountability, and human oversight \cite{act2024eu,regulation2016regulation}. Yet these principles can remain abstract unless they are translated into concrete interaction mechanisms. Contestable AI (CAI) responds to this problem by arguing that people affected by automated decisions should be able to question system outputs, provide counterevidence, seek revision, and receive a meaningful response \cite{leofante2024contestable,freedman_argumentative_2025,cao2026adaptive}. Recent work further suggests that contestability should be designed into AI systems rather than added only as a post-hoc appeal process, because meaningful contestation requires users to see the basis of a recommendation, identify grounds for disagreement, and understand how their challenge changes the outcome \cite{ploug_four_2020,nguyen2026heart2mind}.
Fig.~\ref{fig:fw} illustrates this progress of moving from one-way explanations (from AI to human) to an interactive loop of contestations (between AI and human), wherein the human can request justification, inject corrections, and guide the AI towards more acceptable outcomes.

CoPlan builds on this shift from explainability to contestability. Instead of presenting AI-generated care plans as final recommendations, CoPlan treats them as proposals that can be inspected, challenged, revised, justified, or escalated. This interface-level framing is important because care planning often involves disagreement, uncertainty, and stakeholder negotiation. By combining co-intelligence with contestability, CoPlan aims to support AI-assisted planning while ensuring that clinicians, patients, and caregivers retain meaningful authority over the care plan. In this way, the design goal shifts from automatic plan generation to accountable plan negotiation.

%% file: sec/3_fw.tex
\section{Framework and Interface}
In this section, we present \textit{CoPlan}, a co-intelligent and contestable interface for human-AI care planning through role-based argument graphs. CoPlan supports aging-in-place care planning, where older adults aim to live safely and independently as their needs change. In this setting, care planning requires coordination across clinical evidence, functional needs, psychosocial context, environmental risks, caregiver capacity, and patient preferences.
CoPlan combines domain-specialized AI agents, evidence retrieval, structured argumentation, role-based contestation, and final care plan generation.

The framework is guided by two principles: \textit{co-intelligence} and \textit{contestability}. Co-intelligence positions AI agents and human stakeholders as collaborative contributors with distinct knowledge and responsibilities. Contestability ensures that recommendations remain open to inspection, disagreement, revision, and justification. Therefore, human care planners remain central to the workflow, as they can \textit{accept}, \textit{reject}, \textit{modify}, or \textit{add} arguments before final plan generation. This design reduces manual workload while preserving clinical judgment, human agency, and accountability.\footnote{A video demonstration of CoPlan is available at \url{https://youtu.be/Fel3ZCF3khY}.}

\label{sec:implementation}

\subsection{Problem Environment Formulation} We consider a decision environment in which the goal is to produce a safe and personalized care plan for an older adult living in the community. This task requires synthesizing clinical conditions, functional abilities, environmental risks, and personal preferences. In practice, high-quality care planning requires contributions from several professional roles, as no single discipline can fully capture the complexity of aging-in-place. Our system aims to support this process by creating a structured multi-agent environment in which different professional viewpoints are revealed, compared, and validated.
Formally, we model the environment as a tuple:
\begin{equation}
    \mathcal{M}=\langle P,\mathcal{D},\mathcal{O},\mathcal{A},\Gamma,H,V,\Pi\rangle.
\end{equation}
Here, the patient information $P$ describes health conditions, functional status, and contextual factors. The system retrieves a set of evidence documents $\mathcal{D}$. A generative model uses both $P$ and $\mathcal{D}$ to propose a set of candidate care options $\mathcal{O}$. A care team recruitment mechanism then chooses a subset of healthcare roles, denoted $\mathcal{A}$, which serve as agents providing expert analysis. Each agent produces supporting and challenging arguments for each option, creating an argumentative pool $\Gamma$. A human reviewer may revise this set, producing $\Gamma_H$. A validation operator $V$, grounded in quantitative bipolar argumentation semantics, assigns each argument a degree of acceptability. Finally, a care plan operator $\Pi$ synthesizes a recommended care plan using the weighted argumentative structure.

\begin{figure}[t]
    \includegraphics[width=\linewidth]{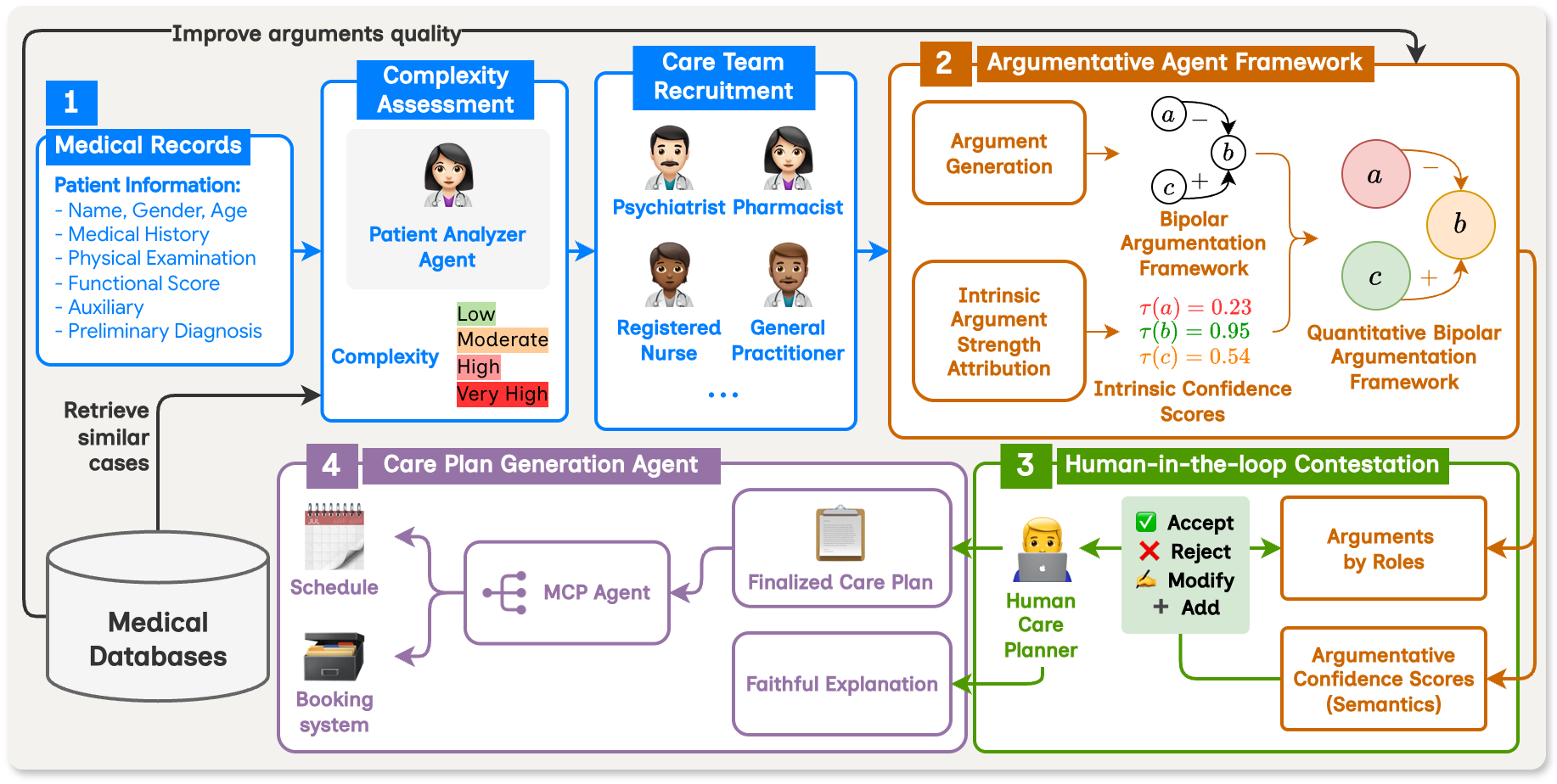} 
    \caption{Our proposed Role-based Contestable Argument Graph Computation Workflow for CoPlan Interface in Aging-in-Place Context.}
    \label{fig:system_overview}
\end{figure}

\subsection{Implementation}
Fig.~\ref{fig:system_overview} shows our proposed framework workflow. It combines an adaptive network of experts with a human-AI collaborative, evidence-based, and contestable care plan generation interface. The result is a comprehensive, human-centered decision-support system that operates across multiple stages.
\subsubsection{Stage 1 – Complexity Assessment and Care Team Recruitment} The system begins by constructing the initial state of the care-planning environment from the client’s demographic profile, medical history, and most recent interRAI~\cite{gray2009sharing} Home Care assessment. The interRAI Home Care assessment is a standardized, comprehensive clinical instrument widely used in community and long-term care settings to capture multidimensional information about an individual’s health status, functional abilities, cognitive condition, and social context. It provides structured, validated measures across domains such as activities of daily living (ADLs), instrumental activities of daily living (IADLs), comorbidities, mental health, and environmental risks. The use of interRAI is motivated by its ability to produce consistent, high-quality, and interoperable patient representations that support care planning, risk stratification, and outcome monitoring across care providers. By grounding the system’s initial state in interRAI-derived data, we ensure that the downstream multi-agent reasoning process operates on a clinically validated and holistic representation of the patient, reducing ambiguity and enabling more reliable coordination among specialized agents.

We denote this structured information as the patient description: $P=\{p_1,\dots,p_k\},$ which includes diagnoses, functional scores, and observed risks. Based on $P$, the system issues targeted retrieval queries across the LLM and the historical medical vector database. This produces a set of relevant documents $\mathcal{D}=\{d_1,\dots,d_n\}$, each annotated with embedding similarity scores and metadata describing source type and reliability.
In parallel, the patient analyzer agent computes a case complexity $c = C(P) \in \{\mathrm{low},\mathrm{moderate},\mathrm{high},\mathrm{very\ high}\},$ based on patterns in comorbidity, psychosocial risk, and functional decline. The score determines the breadth and depth of expertise required. A team recruitment agent then recruits a set of provider agents $\mathcal{A} = S(P,c),$ where each agent $a\in\mathcal{A}$ corresponds to a clinical role (i.e., registered nurse, pharmacist, general practitioner, nutritionist, physical therapist, occupational therapist, psychiatrist, social worker, home health aide, care coordinator), paired with domain-specific prompts and prior expertise. This team forms the basis for the collaborative discussion phase.

In Stage 1, CoPlan uses the LLM to analyze the patient profile and recruit an appropriate interdisciplinary care team. The patient analysis prompt asks the model to extract medical, functional, psychosocial, safety, and coordination needs in a structured JSON format. This output is then used by a second team-selection prompt to choose provider agents whose expertise matches the case complexity.
\definecolor{coplanblue}{HTML}{2596be}

\newtcolorbox{prompttemplate}[1]{
  enhanced,
  breakable,
  colback=gray!3,
  frame hidden,
  boxrule=0pt,
  borderline west={2pt}{0pt}{coplanblue},
  sharp corners,
  left=2.5mm,
  right=2mm,
  top=1mm,
  bottom=1mm,
  before skip=5pt,
  after skip=5pt,
  title=#1,
  fonttitle=\bfseries\small\color{coplanblue},
  coltitle=coplanblue,
  fontupper=\small
}
\begin{prompttemplate}{Patient Analysis Prompt:}
\small
You are a healthcare coordinator analyzing a patient case to determine care needs.

\textbf{Patient Information:} \texttt{<PATIENT\_INFO>}

Analyze this patient and identify: 
(1) primary medical conditions, 
(2) functional needs, 
(3) psychosocial needs, 
(4) care complexity level, 
(5) safety risks, and 
(6) coordination needs.

Return a structured JSON object containing medical conditions, functional status, psychosocial needs, complexity level, safety risks, coordination intensity, and special considerations.
\end{prompttemplate}

\begin{prompttemplate}{Care Team Recruitment Prompt:}
\small
You are assembling an optimal interdisciplinary healthcare team for an elderly patient's care plan.

\textbf{Patient Information:} \texttt{<PATIENT\_INFO>}

\textbf{Patient Analysis:} \texttt{<PATIENT\_ANALYSIS>}

\textbf{Available Healthcare Professionals:} \texttt{<AVAILABLE\_ROLES>}

Return a valid JSON object specifying the selected team, reasoning for each selected role, team size, overall strategy, key collaborations, and potential care gaps.
\end{prompttemplate}

\subsubsection{Stage 2 – Argumentative Agent Framework and Collaborative Discussion} Central to our system are structured discussions and debates between specialized agents that improve factual grounding and reasoning robustness. For each candidate intervention $o_i\in\mathcal{O}$, every provider agent $a_j\in\mathcal{A}$ generates two classes of arguments: one in \textit{support} and one in \textit{against} the intervention. Let $S_{j,i}$ and $C_{j,i}$ denote these two sets, and define the full argumentative pool as: $\Gamma = \bigcup_{i,j} \bigl(S_{j,i} \cup C_{j,i}\bigr).$  
Each argument $x \in \Gamma$ is represented as [content($x$), type($x$), role($x$)], where type($x$) $\in$ \{support, challenge\} and role($x$) $\in \mathcal{A}$ records the provider role that generated it. Agentic RAG is available during this stage, allowing agents to repeatedly query the historical medical vector database for similar case evidence to enhance the factual grounding of their arguments.

\paragraph{Candidate Care Option Generation.}
After retrieving relevant evidence, CoPlan uses the LLM to generate candidate handling options for the care plan. The prompt combines patient information with retrieved medical knowledge and asks for practical options that support independent living, safety, and quality of life.

\begin{prompttemplate}{Care Option Generation Prompt:}
\small
You are an expert geriatric care planner. Based on the following patient information and relevant medical knowledge, generate 2--3 specific handling options for an aging-in-place care plan.

\textbf{Relevant Medical Knowledge:} \texttt{<RAG\_CONTEXT>}

\textbf{Patient Information:} \texttt{<PATIENT\_INFO>}

Provide handling options in the following format:

\texttt{Option 1: [Description]} \\
\texttt{Option 2: [Description]}

Focus on practical and implementable options that support independent living while ensuring safety and quality of life.
\end{prompttemplate}

\paragraph{Quantitative Bipolar Argumentation Framework (QBAF).}
To model the interaction between arguments, we employ QBAF \cite{baroni2019fine,freedman_argumentative_2025}. Let $X = \Gamma$ be the set of all arguments after discussion. We define two relations on $X$: a support relation $R^{+} \subseteq X \times X$ and a challenge relation $R^{-} \subseteq X \times X$. The relation $R^{+}$ captures when one argument supports another. For example, an occupational therapist’s argument that ``the client has frequent bathroom slips'' may support a nurse’s argument that ``installing grab bars will reduce fall risk.'' The relation $R^{-}$ captures conflict, for example, a caregiver’s note that ``the client refuses to use the walker indoors'' may challenge a physiotherapist’s argument that ``a walking program can begin immediately''. Together with a weight function, these elements form the QBAF as:
$\mathcal{Q} = \langle X, R^{+}, R^{-}, \tau \rangle.$
Each argument $x \in X$ receives an intrinsic strength score $\tau(x) \in [0,1]$. This score is computed by a scoring model that evaluates the argument for clinical relevance, its factual consistency with the retrieved documents $\mathcal{D}$, and the transparency of its reasoning. 
Intuitively, $\tau(x)$ measures how convincing the argument is before any interaction with other arguments is taken into account.
The quantitative semantics then computes a final confidence degree $f(x)$ for each argument by combining its intrinsic strength with the influence of supporting and attacking arguments. 
Let $f : X \to [0,1]$ be the degree function we expect to obtain. For each argument $x$, we define an influence term:
\begin{equation}
    I(x,f) = \sum_{y : (y,x)\in R^{+}} \alpha_{y,x} f(y) \quad - \sum_{y :(y,x)\in R^{-}} \beta_{y,x} f(y),
\end{equation}
where $\alpha_{y,x}$ and $\beta_{y,x}$ are non-negative influence weights that control how strongly a supporter or attacker $y$ affects $x$ (all elements in each relation have a pair that contains the target $x$ and the related $y$ argument). The updated degree is given by:
$f(x) = \sigma\bigl(\tau(x) + I(x,f)\bigr),$
where $\sigma$ is a squashing function that keeps the value in $[0,1]$. Starting from the initial vector $f^{(0)}(x) = \tau(x)$, the system iterates this update until convergence, yielding a stable degree $f(x)$ for every argument.
These degrees are then aggregated at the option level. For a given option $o_i$, let $X^{+}_{i}$ be the set of arguments that support $o_i$ and $X^{-}_{i}$ the set that challenge it. We compute an option-level confidence score:
\begin{equation}
    F(o_i) = g\big({f(x) : x \in X^{+}_{i}}, {f(x) : x \in X^{-}_{i}}\big),
\end{equation}
where $g$ is an aggregation function that increases with strong support and decreases with strong challenges. The resulting scores, $F(o_i)$, summarize for each proposed intervention how well it is supported by the multi-agent discussion under the QBAF semantics. These scores are later shown to the human-in-the-loop contestation phase and used by the system when constructing the final care plan.

\paragraph{Role-Based Argument Generation.}
For each selected healthcare agent, CoPlan constructs a role-specific prompt using the agent’s professional background, expertise areas, and care priorities. The agent is then asked to generate both supporting and challenging arguments for each candidate care option. This design ensures that the argumentative pool $\Gamma$ contains not only reasons in favor of a recommendation, but also possible risks, limitations, and conflicts that can later be reviewed by humans.

\begin{prompttemplate}{Role-Based Argument Generation Prompt:}

\small

You are a \texttt{<PROVIDER\_ROLE>} providing expert input on elderly care planning.

\textbf{Expertise Areas:} \texttt{<EXPERTISE\_AREAS>} \\
\textbf{Professional Priorities:} \texttt{<FOCUS\_PRIORITIES>} \\
\textbf{Professional Perspective:} \texttt{<ARGUMENT\_STYLE>}

\textbf{Relevant Medical Knowledge:} \texttt{<RAG\_CONTEXT>}

\textbf{Patient Information:} \texttt{<PATIENT\_INFO>}

\textbf{Care Options Being Evaluated:} \texttt{<CARE\_OPTIONS>}

From your professional perspective, provide 2--3 supporting arguments and 2--3 challenging arguments for each option. Focus on aspects most relevant to your expertise.

Use the following format:

\texttt{Option 1:} \\
\texttt{Support: [support argument]} \\
\texttt{Challenge: [challenge argument]}
\end{prompttemplate}

\begin{figure}[ht]
    \centering
    \shadowbox{\includegraphics[width=\linewidth]{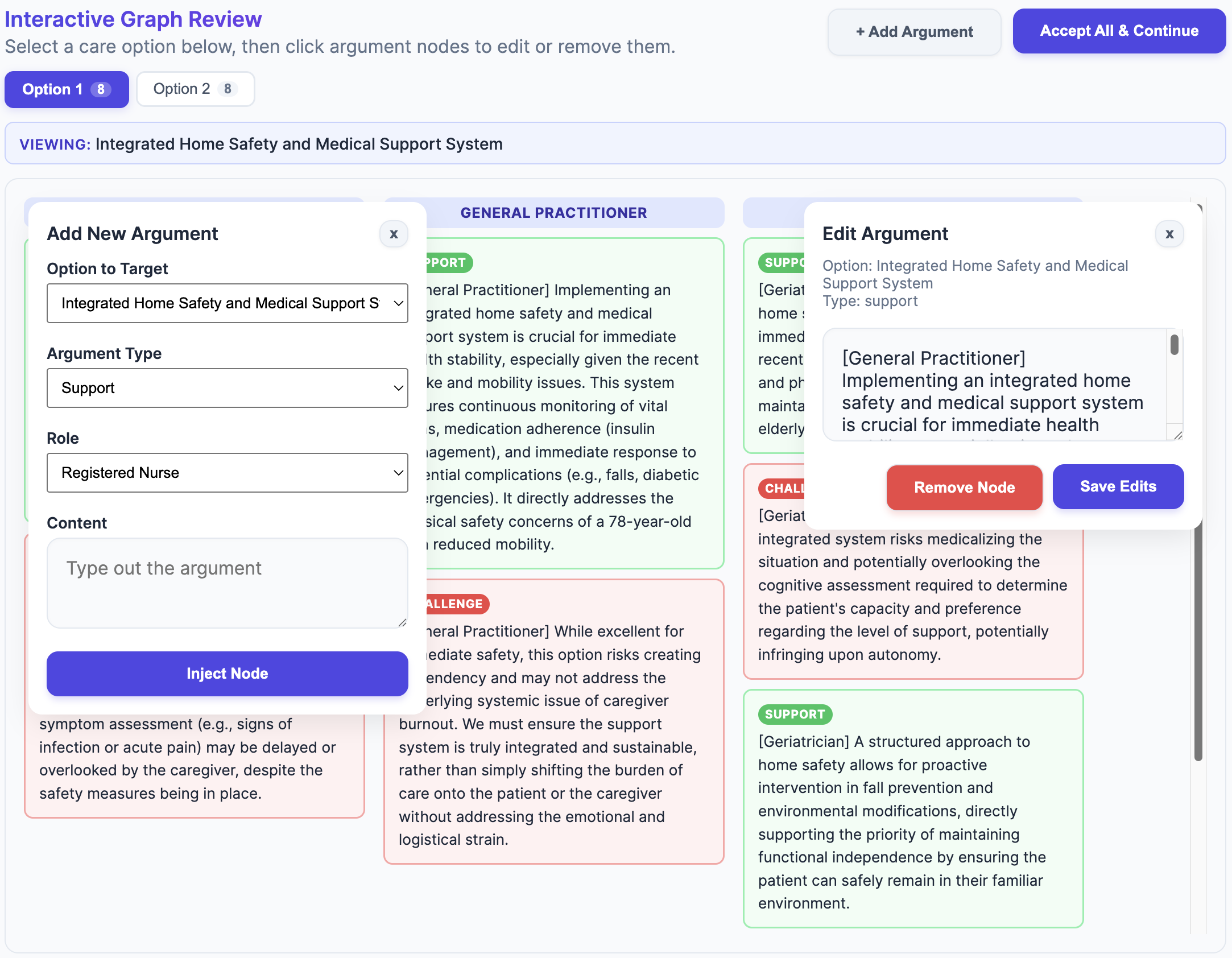}}
    \caption{Interactive Graph Review Panel for role-based human-in-the-loop contestation. The interface allows reviewers to inspect support and challenge arguments, add new arguments, edit existing ones, remove inappropriate nodes, and confirm the revised argumentative structure before proceeding.}
    \label{fig:graphreview}
\end{figure}

\subsubsection{Stage 3 – Role-Based Human-in-the-loop Contestation}

The goal of this stage is to introduce structured contestability while ensuring that clinical judgment remains central to the decision process. The system first generates a participation summary that shows how each agent’s role contributed to the discussion. Because every argument is tagged with its originating role, the system can group arguments by profession and display the distribution of supporting and challenging points raised.

Let $\Gamma$ be the set of all arguments produced in Phase 2. Members of the \textit{human care team} review each argument $x \in \Gamma$, taking into account its intrinsic strength $\tau(x)$, its final degree $f(x)$, and the option-level score $F(o_i)$ for the intervention it relates to. During role-based contestation, team members may: (1) \textbf{accept} an argument, (2) \textbf{reject} an argument judged irrelevant, unsafe, or inconsistent with clinical practice, (3) \textbf{modify} an argument’s content to better reflect the client’s situation, or (4) \textbf{add} new arguments based on professional insight that was not captured by the agent team.

As shown in Fig.~\ref{fig:graphreview}, these actions are supported through an interactive graph review interface. The interface allows users to switch between candidate care options, inspect supporting and challenging arguments grouped by professional role, and directly interact with individual argument nodes. Reviewers can open a form to add a new argument, specify its target option, type, and originating role, or edit an existing argument by revising its content or removing it entirely. This interface operationalizes contestability at the interaction level by making the argumentative structure visible and editable, thereby enabling human reviewers to actively shape the reasoning process rather than merely observe it.

These human edits produce a revised argument set $\Gamma_H \subseteq \Gamma$. 
To maintain internal coherence, the system re-applies the quantitative bipolar semantics to $\Gamma_H$, yielding updated degrees $\Gamma_V = \{(x, f(x)) : x \in \Gamma_H\}$. This update ensures that any human changes propagate through the support $R^{+}$ and challenge $R^{-}$ relations, so that the final scores reflect both the argumentative structure and the reviewer’s expert judgment. After role-based contestation is complete, the \textit{human care planner} reviews the consolidated results, resolves any remaining inconsistencies, and serves as the final authority to validate and approve the argument set before the system proceeds to generate the final care plan.

\paragraph{Argument Validation.}
After human reviewers accept, reject, modify, or add arguments, CoPlan validates the revised argument set using an LLM-based scoring prompt. The model evaluates each argument according to factual accuracy, relevance to elderly care, practical feasibility, and evidence-based reasoning. The resulting scores are used as the initial validity values before being combined with the support and challenge relations in the argument graph.

\begin{prompttemplate}{Argument Validation Prompt}
\small
You are an expert analyst evaluating the validity and relevance of arguments for elderly care planning.

Evaluate the following arguments based on:
(1) factual accuracy, 
(2) relevance to elderly care and aging-in-place, 
(3) practical considerations, and 
(4) evidence-based reasoning.

Provide a validity score between 0 and 1 for each argument, where 0 means completely invalid or irrelevant, 0.5 means moderately valid, and 1 means highly valid and relevant.

\textbf{Arguments to Evaluate:} \texttt{<REVISED\_ARGUMENTS>}

Use the following format:

\texttt{Arg 1 Score: 0.85} \\
\texttt{Arg 2 Score: 0.40}
\end{prompttemplate}


\begin{figure}[htbp]
    \centering
    \shadowbox{\includegraphics[width=\linewidth]{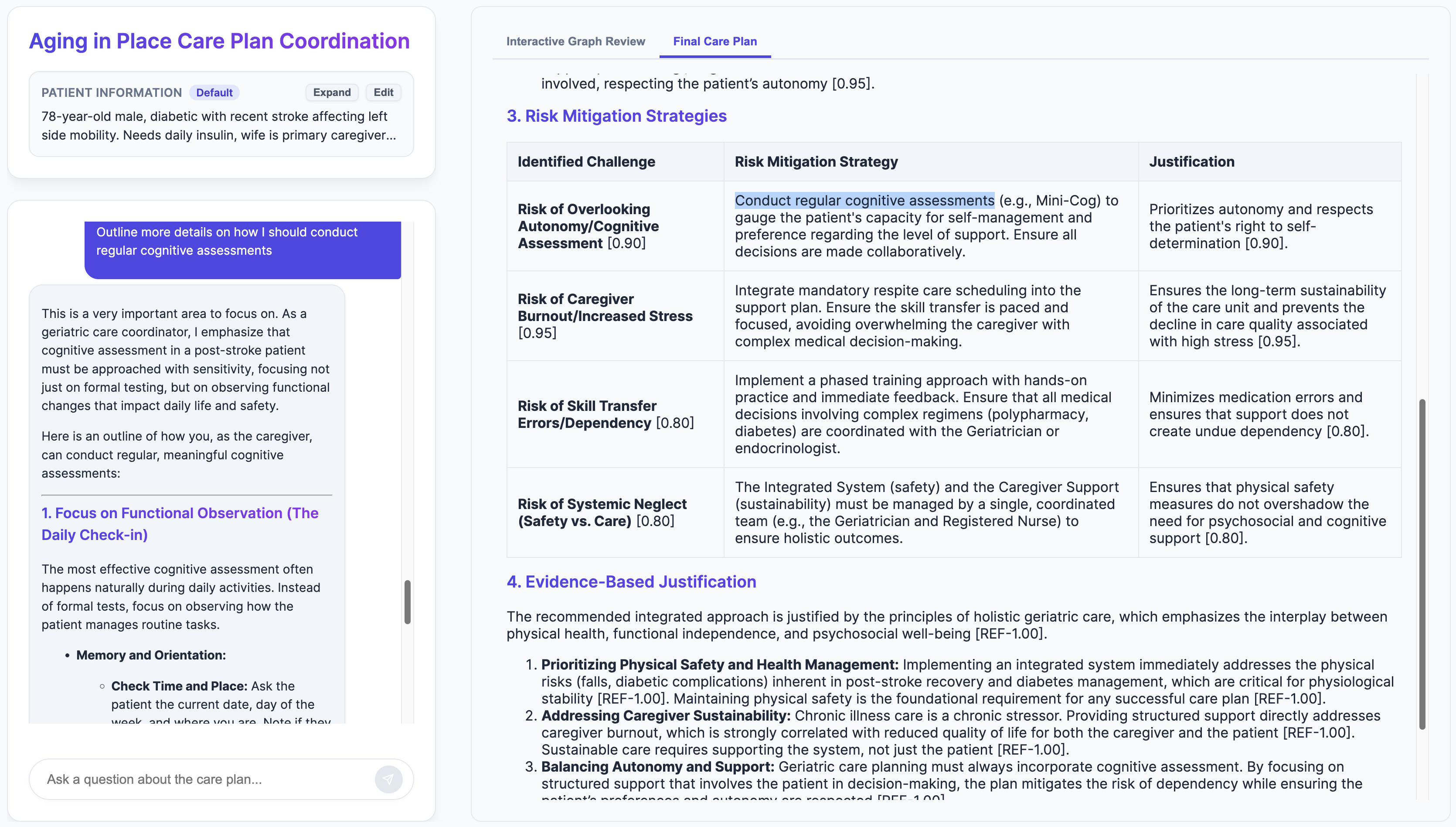}}
    \caption{Final Care Plan Viewer (right panel) and Chatbot Assistant (left panel) for further discussion.}
    \label{fig:finalcareplan}
\end{figure}

\subsubsection{Stage 4 – Final Care Plan Generation and Human Approval}
The care plan generation operator $\Pi$ synthesizes the final plan using the validated arguments $\Gamma_V$, option set $\mathcal{O}$, and retrieved evidence $\mathcal{D}$. For each intervention $o_i$, the operator uses its updated option-level score $F(o_i)$ to determine its priority, recommendation strength, and required risk-mitigation steps. The resulting care plan is a prescriptive, evidence-based set of recommendations tailored to the client’s physical, psychosocial, and environmental context.

CoPlan uses the validated arguments and retrieved evidence to generate the final care plan. The prompt explicitly asks the LLM to consider argument validity scores, cite retrieved evidence using reference identifiers, and organize the output into prioritized recommendations, implementation steps, risk mitigation strategies, and evidence-based justifications.

\begin{prompttemplate}{Final Care Plan Generation Prompt}
\small

You are an expert geriatric care planner. Based on the validated arguments and medical knowledge, create a comprehensive revised care plan for the elderly patient.

When using information from the medical knowledge provided, cite it using \texttt{[REF-X]} format.

\textbf{Relevant Medical Knowledge:} \texttt{<RAG\_CONTEXT>}

\textbf{Patient Information:} \texttt{<PATIENT\_INFO>}

\textbf{Original Handling Options with Validated Arguments:} \texttt{<VALIDATED\_ARGUMENTS>}

Based on the arguments, their validity scores, and the medical knowledge provided, provide:
(1) a prioritized list of recommended handling options,
(2) specific implementation steps,
(3) risk mitigation strategies for identified concerns, and
(4) evidence-based justification with references.

Consider the strength of arguments when making recommendations.
\end{prompttemplate}

As shown in Fig.~\ref{fig:finalcareplan}, the generated plan is presented through a final care plan viewer that organizes recommendations into structured sections, including risk mitigation strategies and evidence-based justification. Each recommendation is linked to an identified challenge, a proposed mitigation strategy, and a justification with confidence information, making the final plan easier to inspect and verify. The interface also includes a chatbot assistant in the left panel, allowing users to ask follow-up questions, request clarification, or discuss how specific recommendations should be implemented. This design supports the transition from validated argumentation to actionable care planning while keeping the plan open to further human review and discussion.

Importantly, the generated care plan is not treated as an automatically finalized decision. After Stage 4, the human care planner remains responsible for reviewing the proposed plan, identifying infeasible, contradictory, unsafe, or resource-incompatible recommendations, and requesting revision before implementation. If a recommendation cannot be realized because required services, personnel, funding, caregiver capacity, or local resources are unavailable, the care planner may reopen the relevant arguments, add feasibility constraints, revise the recommendation, or reject the plan component.

\paragraph{Model Context Protocol (MCP) Agent.} 
After the final plan is produced, the system moves from analysis to practical follow-up. Virtual agents act on behalf of real-world providers and implement the recommended interventions generated by $\Pi$ by translating them into concrete scheduling tasks. This is achieved through the MCP \cite{protocol_2024}, which provides the booking agent with the necessary scheduling functions. With this integration, the system can arrange appointments, request assessments, and schedule provider visits, ensuring that the care plan is carried out in a timely and organized manner.


%% file: sec/4_res.tex
\section{Results}
\begin{figure}[b!]
    \centering
    \includegraphics[width=\linewidth]{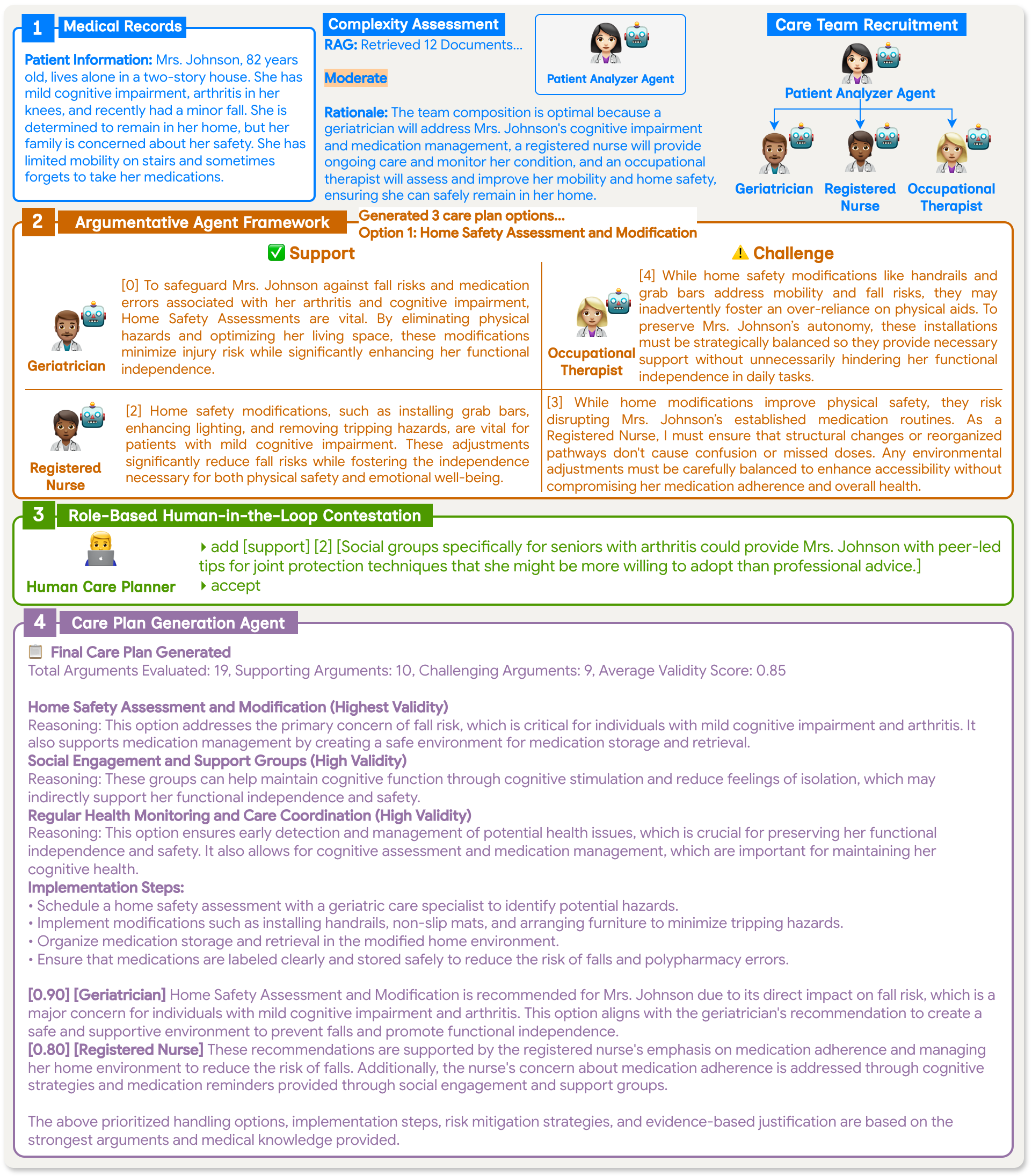}
    \caption{Demonstrative result of CoPlan in an aging-in-place care planning scenario.}
    \label{fig:example}
\end{figure}

We present a demonstrative result of CoPlan in an aging-in-place care planning scenario. Fig.~\ref{fig:example} illustrates the end-to-end pipeline and shows how the system moves from patient information to a finalized care plan through four connected stages. Starting from the client’s medical and contextual records, the system first performs complexity assessment and care team recruitment, identifying the most relevant provider roles for the case. It then generates candidate care options and collects supporting and challenging arguments from specialized agents, making the reasoning process explicit rather than hidden.

This result also highlights the role of human oversight in the workflow. In Stage 3, the human care planner reviews the agent-generated arguments and can accept, revise, or add points based on professional judgment. This contestation process ensures that the final care plan is not simply the output of automated reasoning but the result of a co-intelligent, contestable collaboration between human stakeholders and AI agents.

The final stage demonstrates how the validated arguments are synthesized into a structured care plan. In the illustrated case, the system prioritizes home safety assessment and modification, social engagement and support groups, and regular health monitoring and care coordination as the highest-validity recommendations. The final output also includes implementation steps, argument validity summaries, and evidence-based justification, showing how CoPlan supports transparent, actionable, and human-centered care planning.

%% file: sec/5_conc.tex
\section{Conclusion}
This paper presented \textit{CoPlan—a Co-Intelligent and Contestable Interface for Human-AI Care Planning}. CoPlan treats AI recommendations as reviewable proposals rather than fixed decisions. It analyzes patient information, recruits relevant care roles, generates interventions and supporting or opposing arguments, and allows care planners to accept, reject, revise, or add arguments before producing the final plan.
The aging-in-place scenario illustrates how CoPlan combines patient context, role-based expertise, structured contestation, and evidence-based recommendations. Its visible, editable, and auditable reasoning supports co-intelligence while preserving human authority over care decisions.
A current limitation is that validity and confidence scores support review rather than autonomous decision-making. Future work should evaluate alternative aggregation methods and examine their effects on transparency, fairness, legitimacy, and overreliance. Studies should also assess whether clinicians, patients, and caregivers can effectively express value-based concerns through contestation.
Overall, trustworthy AI-supported care planning requires not only accurate recommendations or explanations, but also accountable negotiation. CoPlan provides a design direction that preserves human agency, clinical responsibility, and meaningful oversight.

%% file: sec/6_ack.tex
\section{Acknowledgment}
\label{sec:ack}
This work is supported by NSERC Discovery Grant No RGPIN-2025-04478 and NSERC Discovery Supplement Award No DGECR-2025-00129. It is also funded by National Research Council Canada Aging in Place Challenge grant AiP-301-1 D-CGA@home.

\section{Disclosure of Interests}
The authors have no competing interests to declare that are relevant to the content of this article.